\PassOptionsToPackage{table,xcdraw}{xcolor}
\documentclass[11pt]{article}
\usepackage[final]{acl}

\usepackage{times}
\usepackage{latexsym}
\usepackage{times}
\usepackage{latexsym}
\usepackage{amsmath}
\usepackage{multirow}
\usepackage{longtable}
\usepackage{algorithm}
\usepackage{algpseudocode}
\usepackage{siunitx}
\usepackage{tabularx}
\usepackage{makecell} 
\usepackage{float}
\usepackage{longtable}
\usepackage{authblk}
\usepackage{graphicx}
\usepackage{booktabs}
\usepackage{adjustbox}
\usepackage[T1]{fontenc}
\usepackage[utf8]{inputenc}
\usepackage{microtype}
\usepackage{inconsolata}

\title{See What You Need: Query-Aware Visual Intelligence through Reasoning-Perception Loops}

\author{
 \textbf{Zixuan Dong\textsuperscript{1}},
 \textbf{Baoyun Peng\textsuperscript{1,\thanks{Corresponding Author}}},
 \textbf{Yufei Wang\textsuperscript{1}},
 \textbf{Lin Liu\textsuperscript{1}},
 \textbf{Xinxin Dong\textsuperscript{1}},
 \\
 \textbf{Yunlong Cao\textsuperscript{1}},
 \textbf{Xiaodong Wang\textsuperscript{1,\textsuperscript{*}}}
 
\\
\\
 \textsuperscript{1}{College of Computer, National University of Defense Technology}
\\
}

\begin{document}
\maketitle
\begin{abstract}  
Human video comprehension demonstrates dynamic coordination between reasoning and visual attention, adaptively focusing on query-relevant details. 
However, current long-form video question answering systems employ rigid pipelines that decouple reasoning from perception, leading to either information loss through premature visual abstraction or computational inefficiency through exhaustive processing. 
The core limitation lies in the inability to adapt visual extraction to specific reasoning requirements—different queries demand fundamentally different visual evidence from the same video content. 
In this work, we present CAVIA, a training-free framework that revolutionizes video understanding through reasoning-perception coordination. Unlike conventional approaches where visual processing operates independently of reasoning, CAVIA creates a closed-loop system where reasoning continuously guides visual extraction based on identified information gaps. 
CAVIA introduces three innovations: (1) hierarchical reasoning-guided localization to precise frames; (2) cross-modal semantic bridging for targeted extraction; (3) confidence-driven iterative synthesis.
CAVIA achieves state-of-the-art performance on challenging benchmarks: EgoSchema (65.7\%, +5.3\%), NExT-QA (76.1\%, +2.6\%), and IntentQA (73.8\%, +6.9\%), demonstrating that dynamic reasoning-perception coordination provides a scalable paradigm for video understanding.
\end{abstract}

\begin{figure}[t]
  \centering
  \includegraphics[width=0.92\linewidth]{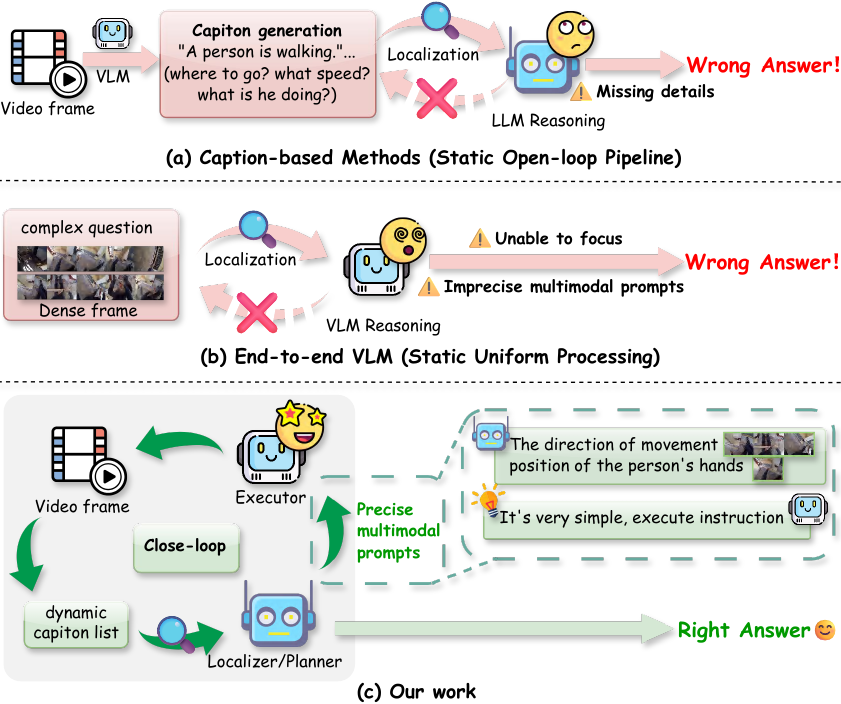} 
    \caption{CAVIA addresses the limitations of static caption-based methods and unfocused VLMs through closed-loop coordination, dynamically generating precise multimodal prompts to extract query-specific spatiotemporal details for accurate video understanding.}
  \label{fig:motivation}
\end{figure}

\section{Introduction}

Video question answering (VideoQA) entails the comprehension of intricate temporal dynamics, spatial relationships, and semantic content within video sequences~\cite{xiao2022video}. While substantial advancements have been made in analyzing short video clips, long-form video understanding presents unique challenges~\cite{wu2024longvideobench}: the sparse distribution of query-relevant information across extensive sequences and the computational burden of processing lengthy content while preserving fine-grained details~\cite{song2024moviechat,li2023discovering}. Given a natural language query and an extended video, systems must efficiently identify pertinent temporal segments while maintaining a holistic understanding of the video content~\cite{fei2024video}.


Current methods achieve significant progress in addressing this challenge, typically through two predominant paradigms. 
Caption-based approaches~\cite{snapcap2025} convert visual information into textual descriptions, leverage the powerful reasoning capabilities of language models for answer generation. Conversely, end-to-end vision-language models~\cite{youk2024fma} achieve comprehensive visual understanding by retaining and processing raw visual data uniformly. 
However, both paradigms exhibit a critical limitation stemming from the decoupling of reasoning and perception~\cite{wang2024stair,ko2023large}: perception passively feeds information to reasoning modules without adaptive feedback, preventing cognition from guiding perception.
Consequently, captioning methods inevitably suffer from information loss due to premature abstraction of visual details, while end-to-end models incur computational overload from exhaustive, indiscriminate processing of all visual content~\cite{shen2023accurate}.

Human cognition demonstrates a dynamic interplay between reasoning and visual attention, where perceptual focus continuously adapts based on evolving reasoning states~\cite{santos2025infinityvideo}, revealing a radically different architecture where reasoning and perception exist in constant dialogue, mutually informing and reshaping each other.
Consider two distinct queries: answering ``what tool was used?'' demands detailed spatial analysis of objects and their interactions, while determining ``why did the person stop?'' necessitates understanding temporal sequences and broader contextual cues. 
This synergistic guidance of perception by higher-level cognition starkly contrasts with existing computational methods' fundamental limitation: the decoupling of reasoning and perception. Figure~\ref{fig:motivation} illustrates these limitations and motivates our closed-loop approach. 

Inspired by the profound interplay of reasoning and perception inherent to human cognition, we present \textbf{CAVIA} (\textbf{C}losed-loop \textbf{A}daptive \textbf{V}ideo \textbf{I}ntelligence \textbf{A}gent), which fundamentally rethinks video understanding by establishing dynamic coordination between reasoning and perception. Our key insight: effective comprehension requires neither exhaustive processing nor static captions, but adaptive mechanisms where reasoning continuously guides visual extraction---mirroring human cognitive processes. Unlike prior iterative approaches that perform simple caption corrections \cite{zhang2025vl}, CAVIA introduces a sophisticated feedback loop that dynamically enriches visual descriptions based on identified reasoning gaps. CAVIA introduces three technical innovations:
\begin{itemize}
\setlength{\itemsep}{-3pt}
\item \textbf{Hierarchical Reasoning-Guided Localization}: Progressively narrows from caption clusters to precise frames through LLM-guided analysis.
\item \textbf{Cross-Modal Semantic Bridging}: Translates reasoning gaps into targeted visual extraction directives for spatial-temporal analysis.
\item \textbf{Confidence-Driven Iterative Synthesis}: Orchestrates progressive refinement through reasoning-perception cycles, with convergence determined by confidence metrics.
\end{itemize}
By coordinating pretrained VLMs and LLMs in a training-free framework, CAVIA achieves state-of-the-art performance: EgoSchema (65.7\%, +5.3\%), NExT-QA (76.1\%, +2.6\%), and IntentQA (73.8\%, +6.9\%). These results validate dynamic reasoning-perception coordination as a scalable paradigm for long-form video understanding







\begin{figure*}[t]
  \centering
  \includegraphics[width=\textwidth]{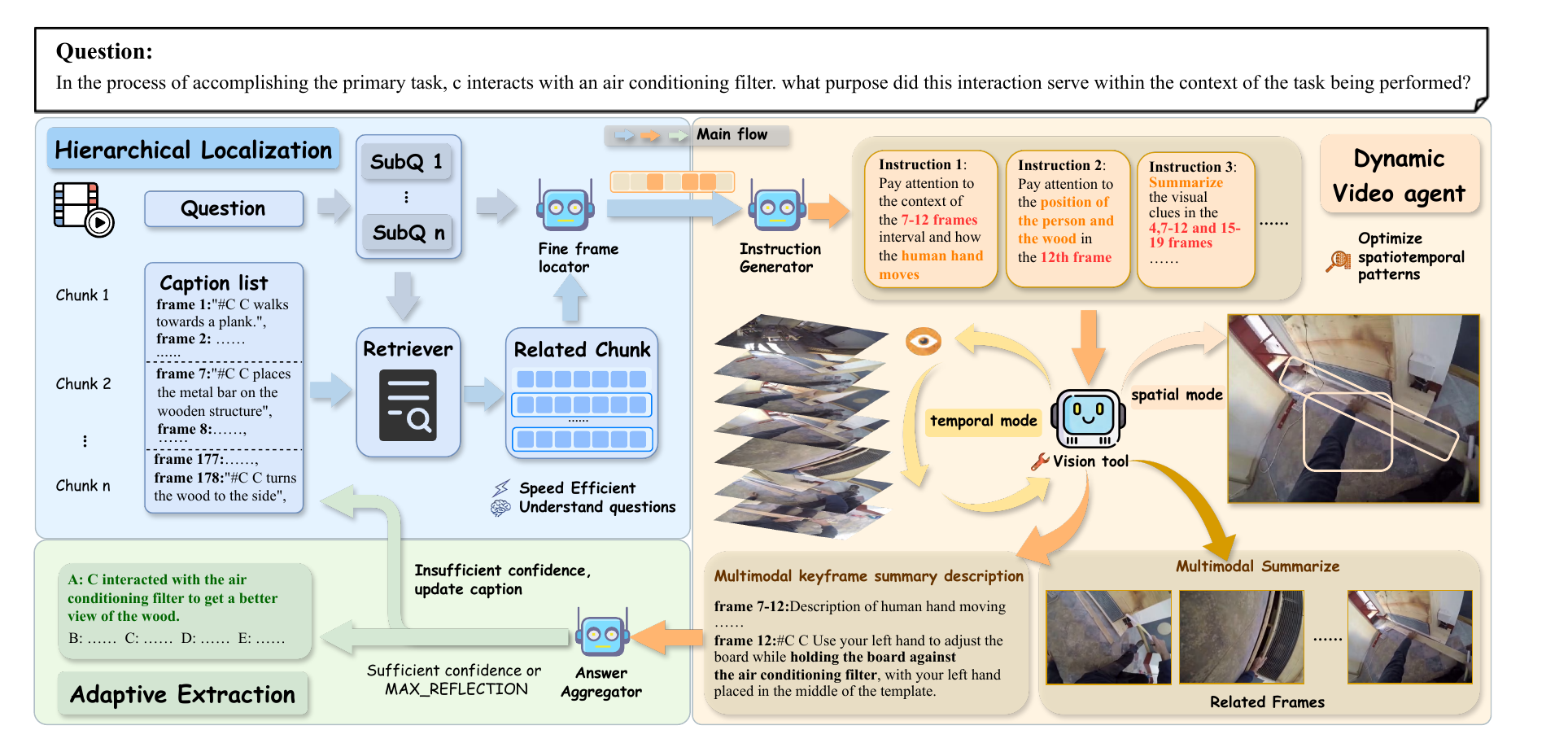} 
      \caption{The CAVIA framework architecture. The system operates through closed-loop coordination between three components: Hierarchical Localization (blue) progressively narrows from caption chunks to specific frames via retrieval and LLM analysis; Dynamic Prompting (orange) generates targeted spatial-temporal instructions based on identified information gaps; Adaptive Extraction (green) employs VLMs in temporal or spatial mode to extract query-specific visual details. The iterative process continues until sufficient confidence is achieved.}
  \label{fig:pipeline}
\end{figure*}

\section{Related Work}
\subsection{Long-form Video Question Answering}
Early VideoQA methods with CNN-RNN architectures faced context window limitations \cite{sharma2022convolutional}. Transformer-based approaches improved temporal modeling \cite{chen2024spatio} but suffer from quadratic complexity. Recent works address efficiency through hierarchical processing: OptiGQA \cite{10.1007/978-981-96-9961-2_17} and VideoTree \cite{wang2025videotree} use multi-level structures, while MIST \cite{gao2023mist} decomposes spatial-temporal attention into cascaded selection modules. For extremely long videos, \cite{nguyen2024encoding} integrates State Space Layers for global semantic encoding. However, these \textit{static pipelines} process all queries uniformly regardless of complexity.

\subsection{Multimodal Architectures for Video Understanding}
Video understanding systems coordinate LLMs and VLMs through various architectures \cite{10438044, jin2024efficient}. Recent efficiency improvements include sparse memory (MovieChat \cite{song2024moviechat}, VideoINSTA \cite{liao2024videoinsta}), streaming processing (Flash-VStream \cite{zhang2024flash}), and memory-based reasoning (Glance-Focus \cite{bai2023glance}). SeViLA \cite{yu2023sevila} leverages image-language models with dual-module chaining for keyframe localization. While DrVideo \cite{ma2025drvideo} introduces iterative refinement, it lacks \textit{fine-grained temporal localization} and \textit{targeted multimodal prompting}, operating at coarse segments without precise spatial-temporal instructions.

\subsection{Adaptive Reasoning and Cross-Modal Coordination}
Adaptive frameworks implement query-specific processing through predefined selection (VideoAgent \cite{10.1007/978-3-031-72989-8_4}), text-visual alignment (LeAdQA \cite{dong2025leadqallmdrivencontextawaretemporal}), or LLM-based temporal reasoning (LLaMA-VQA \cite{ko2023llama}). However, these approaches operate via \textit{open-loop reasoning}—once visual features are extracted, models cannot request targeted information to address reasoning gaps, precluding iterative refinement crucial for complex video understanding.

\section{Method}
We present CAVIA, a closed-loop framework for long-form video question answering that establishes dynamic coordination between reasoning and perception. Unlike static pipelines that process videos uniformly, CAVIA introduces query-aware mechanisms where reasoning continuously guides visual extraction.

\subsection{Problem Formulation and Framework Overview}
Given a video $V$ with uniformly sampled frames $\{f_1, f_2, \ldots, f_T\}$ and corresponding captions $\{c_1, c_2, \ldots, c_T\}$, we answer question $q$ by dynamically identifying and extracting query-specific visual details rather than exhaustively processing all information.

As illustrated in Figure \ref{fig:pipeline}, CAVIA operates through four interconnected components forming a closed-loop system: (1) \textbf{Coarse-to-Fine Temporal Localization} that hierarchically narrows from caption blocks to specific frames, (2) \textbf{Reasoning Gap Identification} where LLMs detect missing information in current context, (3) \textbf{Targeted Multimodal Prompting} that generates precise spatial-temporal instructions for VLMs, and (4) \textbf{Iterative Caption Enhancement} that progressively enriches descriptions based on extracted details. This closed-loop architecture enables reasoning to actively guide perception throughout the process.

\subsection{Coarse-to-Fine Temporal Localization}
We introduce a coarse-to-fine strategy that adapts processing granularity to query complexity. For the initial localization, we partition the caption sequence into semantically coherent blocks and employ a retrieval mechanism that considers both the original query and decomposed sub-questions. This multi-granular matching captures both holistic relevance and specific detail alignment:
\[
\mathcal{C}_{\text{rel}} = \underset{C}{\arg\max} \, \alpha S(q, C) + \beta \sum_{q_i \in D(q)} S(q_i, C)
\]
where $S(q, C)$ measures semantic similarity between query $q$ and caption block $C$, $D(q)$ represents decomposed sub-questions, and $\alpha, \beta$ are weighting parameters. This formulation captures both holistic relevance and specific detail alignment. Within selected caption blocks, our Fine Frame Locator employs LLM analysis to identify frames containing critical visual information. Unlike methods that uniformly sample frames, this query-driven selection ensures computational resources concentrate on discriminative content. The LLM evaluates each caption within relevant blocks, identifying those that potentially contain answer-critical information but lack sufficient detail.

\subsection{Targeted Multimodal Prompting}
The core innovation of CAVIA lies in our iterative enhancement mechanism that dynamically enriches visual understanding based on reasoning requirements. When the LLM detects information gaps in the current context, it generates targeted instructions that guide the VLM to extract specific missing details. The visual tool is implemented using Qwen2.5-VL-7B\footnote{Available at \url{https://huggingface.co/Qwen/Qwen2.5-VL-7B-Instruct}}\cite{bai2025qwen2}, which provides strong visual-language understanding capabilities for both temporal and spatial analysis.

Our Instruction Generator operates in two complementary modes based on the nature of identified gaps:

\textbf{Temporal Enhancement Mode:} When reasoning requires understanding of actions, movements, or temporal transitions, the system selects coherent frame sequences that capture motion dynamics. Unlike static frame sampling, we adaptively determine the temporal span based on action complexity:
\[
I_{\text{temp}} = \underset{t_s, t_e}{\arg\max} \, \text{AC}(f_{t_s:t_e}) \cdot \text{QR}(q, f_{t_s:t_e})
\]
where $\text{AC}(f_{t_s:t_e})$ measures action coherence across frames $t_s$ to $t_e$, and $\text{QR}(q, f_{t_s:t_e})$ evaluates query relevance. This ensures we capture complete action sequences rather than isolated frames. The VLM then processes these intervals with temporal-aware prompts, generating descriptions that capture motion patterns, action sequences, and temporal relationships absent from initial captions.

\textbf{Spatial Enhancement Mode:} For queries requiring fine-grained visual details about objects, scenes, or spatial relationships, prompts such as ``Focus on the tool in the person's right hand'' guide VLMs to extract fine-grained visual attributes from specific regions, ensuring focused analysis rather than generic description.

Algorithm \ref{alg:visual_tool_invocation} details this instruction-driven visual processing mechanism, showing how targeted prompts are generated based on identified reasoning gaps. This targeted enhancement mechanism represents a fundamental departure from existing approaches. Rather than relying on pre-computed visual features or exhaustive frame processing, we dynamically generate visual queries that address specific reasoning gaps, ensuring efficient yet comprehensive understanding.

\begin{algorithm}[ht]
\caption{Instruction-Driven Visual Processing}
\label{alg:visual_tool_invocation}
\begin{algorithmic}[1]
\Require Question $q$, relevant caption blocks $\mathcal{C}_{\text{rel}}$, video frames $\mathcal{F}$, frame-caption mapping $\mathcal{M}$
\Ensure Refined context $\mathcal{C}_{\text{ref}}$, final answer $\hat{y}$
\State $\mathcal{C}_{\text{ref}} \gets \emptyset$, $\mathcal{I} \gets \textsc{GenInst}(q, \mathcal{C}_{\text{rel}})$
\For{each instruction $i \in \mathcal{I}$}
    \If{\textsc{IsTemp}(i)}
        \State $I_t \gets \textsc{SelTime}(\mathcal{F}, i)$
        \State $V_t \gets \textsc{FeatTime}(I_t)$
        \State $\mathcal{C}_{\text{ref}} \gets \mathcal{C}_{\text{ref}} \cup \textsc{DescSeq}(V_t, i)$
    \ElsIf{\textsc{IsSpat}(i)}
        \State $f \gets \textsc{LocFrame}(\mathcal{F}, i)$
        \State $V_s \gets \textsc{FeatObj}(f, i)$
        \State $\mathcal{C}_{\text{ref}} \gets \mathcal{C}_{\text{ref}} \cup \textsc{DescReg}(V_s, i)$
    \EndIf
\EndFor
\State $\mathcal{D} \gets \textsc{Synth}(\mathcal{C}_{\text{ref}}, q)$
\State $\hat{y}_L \gets \textsc{LLMPred}(q, \mathcal{D})$
\State $\hat{y}_M \gets \textsc{VLMPred}(q, I_t, f)$
\State $\hat{y} \gets \textsc{Fuse}(\hat{y}_L, \hat{y}_M)$
\If{\textsc{LowConf}($\hat{y}$)}
    \State $\mathcal{C}_{\text{ref}} \gets \textsc{UpdateCap}(\mathcal{C}_{\text{ref}}, I_t, f)$
    \State $\hat{y} \gets \Call{RefineAnswer}{q, \mathcal{C}_{\text{ref}}, \mathcal{F}, \mathcal{M}}$
\EndIf
\State \Return $\mathcal{C}_{\text{ref}}, \hat{y}$
\end{algorithmic}
\end{algorithm}

\subsection{Iterative Caption Enhancement and Cross-Modal Synthesis}

The final component orchestrates LLM reasoning and VLM perception through confidence-based iterative refinement. Unlike single-pass systems, our framework continues refinement until sufficient confidence is achieved or computational budget is exhausted, enabling progressive improvement through query-specific processing.

In the synthesis stage, the system transcends simple prediction fusion by leveraging the LLM's linguistic reasoning capabilities in conjunction with the VLM's visual understanding. The VLM generates descriptive summaries based on questions and visual content, providing contextual references that enhance the LLM's judgment process.

The answer fusion mechanism adaptively weights contributions from both modalities based on query characteristics and confidence metrics:
\[
\hat{y} = \arg\max_y \, \alpha P_L(y|q, \mathcal{C}^+) + (1-\alpha) P_V(y|q, \mathcal{F})
\]

where $P_L$ and $P_V$ represent probability distributions from LLM and VLM respectively, $\mathcal{C}^+$ denotes enhanced captions, and $\alpha$ dynamically adjusts based on query type and information completeness. This adaptive fusion ensures each modality contributes according to its strengths.

When available information remains insufficient, the framework initiates iterative refinement by updating captions to incorporate missing details. During updates, the system integrates temporal sequence information and previously overlooked visual details. The VLM dynamically adjusts frame sequence processing, incorporating both temporal context through frame intervals and spatial details based on enriched captions.

The iterative nature creates a virtuous cycle: initial coarse understanding guides targeted information extraction, which enables more precise reasoning. Each iteration refines the context through query-specific processing, transforming video understanding from exhaustive computation to intelligent, adaptive extraction.

\begin{table*}[ht]
\centering
\setlength{\tabcolsep}{5pt} 
\begin{tabular}{lcccccccc}
\toprule
\multirow{2}{*}{\textbf{Method}} & \multirow{2}{*}{\textbf{LLM}} & 
\multicolumn{2}{c}{\textbf{EgoSchema}} & 
\multicolumn{4}{c}{\textbf{NExT-QA}} & 
\multicolumn{1}{c}{\textbf{IntentQA}} \\
\cmidrule(lr){3-4} \cmidrule(lr){5-8} \cmidrule(l){9-9} 
& & \textbf{Sub.} & \textbf{Full} & 
\textbf{Tem.} & \textbf{Cau.} & \textbf{Des.} & \textbf{Avg.} & 
\textbf{Avg.} \\
\midrule
\rowcolor[HTML]{D3D3D3} 
\multicolumn{9}{c}{\textbf{Vision-Centric Multimodal Architectures}} \\ 
SeViLA \cite{yu2023self} & GPT-3.5 & - & - & 61.3 & 61.5 & 75.6 & 63.6 & 60.9 \\
LongViViT \cite{papalampidi2024simple} & GPT-3.5 & 56.8 & 33.3 & - & - & - & - & - \\
MC-ViT-L \cite{balavzevic2024memory} & GPT-3.5 & 65.8 & 52.9 & - & - & - & - & - \\
InternVideo2 \cite{wang2024internvideo2} & GPT-4 & 66.4 & 55.8 & - & - & - & - & - \\
\midrule
\rowcolor[HTML]{D3D3D3} 
\multicolumn{9}{c}{\textbf{LLM-Driven Cross-Modal Synergy}} \\ 
MVU \cite{ranasinghe2024understanding} & GPT-4 & 60.3 & 37.6 & 55.4 & 48.1 & 64.1 & 55.2 & - \\
LLoVi \cite{zhang2024simple} & GPT-4 & 57.6 & 50.3 & 61.0 & 69.5 & 75.6 & 67.7 & 64.0 \\
VideoAgent \cite{10.1007/978-3-031-72989-8_4} & GPT-4 & 60.2 & 54.1 & 64.5 & 72.7 & 81.1 & 71.3 & - \\
MoReVQA \cite{min2024morevqa} & GPT-3.5 & - & - & 64.6 & 70.2 & - & - & 69.2 \\
OptiGQA \cite{10.1007/978-981-96-9961-2_17} & GPT-4 & - & - & 65.8 & 72.3 & 78.0 & 71.0  & 68.7 \\
VideoTree \cite{wang2025videotree} & GPT-4 & 66.2 & 61.1 & 67.0 & 75.2 & \underline{81.3} & 73.5 & 66.9 \\
\midrule
\textbf{CAVIA(ours)} & GPT-4 & 70.8 & \textbf{65.7} & 70.2 & 75.7 & 79.2 & 74.5 & 72.8 \\
\textbf{CAVIA(ours)} & GPT-4.1 & \textbf{71.6} & 65.3 & \textbf{71.0} & \textbf{78.2} & 79.6 & \textbf{76.1} & \textbf{73.8} \\
\bottomrule
\end{tabular}
\caption{Performance comparison on EgoSchema, NExT-QA, and IntentQA benchmarks. Methods are categorized into Vision-Centric Multimodal Architectures and LLM-Driven Cross-Modal Synergy paradigms. For NExT-QA, we report results on Temporal (Tem), Causal (Cau), and Descriptive (Des) question types. Bold indicates best performance; underlined indicates second best; "-" denotes results not reported in original papers.} 
\label{tab:comparison}
\end{table*}

\section{Experiments}
We conduct comprehensive experiments to validate CAVIA's effectiveness in long-form video question answering. Our evaluation addresses three key questions: (1) Does closed-loop reasoning-perception coordination outperform static pipelines? (2) How does dynamic prompt generation improve over fixed visual processing? (3) What are the contributions of each component in our iterative framework? We evaluate on three challenging benchmarks---EgoSchema, NExT-QA, and IntentQA---demonstrating that CAVIA achieves state-of-the-art performance while providing insights into the importance of adaptive visual information extraction.

\subsection{Datasets and Evaluation Metrics}
We evaluate CAVIA on three complementary datasets that test different aspects of video understanding:

\textbf{EgoSchema}\cite{mangalam2023egoschema}: Contains 250+ hours of first-person videos with 5,000 multiple-choice questions. Provides both Subset and Fullset evaluations, covering diverse human activities and human-object interactions.

\textbf{NExT-QA}\cite{xiao2021next}: Comprises 5,440 videos (~44 seconds each) with Temporal (Tem.), Causal (Cau.), and Descriptive (Des.) question types, emphasizing temporal and causal reasoning evaluation.

\textbf{IntentQA}\cite{li2023intentqa}: Focuses on intent-based reasoning requiring understanding of causal-temporal relationships and underlying motivations in sequential activities.

We use accuracy as the primary evaluation metric, reporting the percentage of correctly answered multiple-choice questions. For EgoSchema, we evaluate on both Subset and Fullset categories; for NExT-QA, we report Temporal, Causal, Descriptive, and Average performance; for IntentQA, we report overall average accuracy.

\subsection{Experimental Results}
\subsubsection{Comparison with State-of-the-Art Methods}
Table \ref{tab:comparison} presents comprehensive comparisons with existing approaches, categorized into two paradigms: Vision-Centric Multimodal Architectures that primarily rely on visual features, and LLM-Driven Cross-Modal Synergy methods that leverage language models for reasoning.

\textbf{Overall Performance:} CAVIA achieves new state-of-the-art results across all benchmarks. On EgoSchema, we attain 71.6\% accuracy on Subset (+5.4\% over VideoTree) and 65.7\% on Fullset (+4.6\%), demonstrating significant improvements in long-form video understanding. For NExT-QA, CAVIA reaches 76.1\% average accuracy, surpassing the previous best by 2.6\%. Most notably, on the challenging IntentQA dataset, our method achieves 73.8\%, a substantial 4.6\% improvement over MoReVQA.

\textbf{Analysis by Question Type:} The performance breakdown reveals CAVIA's strengths in complex reasoning tasks. For temporal questions on NExT-QA, we achieve 71.0\% accuracy compared to VideoTree's 67.0\%, demonstrating the effectiveness of our coarse-to-fine localization strategy. In causal reasoning (78.2\% vs. 75.2\%), the gains highlight how targeted multimodal prompting extracts crucial cause-effect information missed by static approaches. While VideoTree slightly outperforms on descriptive questions (81.3\% vs. 79.6\%), CAVIA's consistent superiority in temporal and causal categories—which require deeper understanding—validates our focus on dynamic information extraction.

\textbf{Vision-Centric vs. LLM-Driven Approaches:} The results clearly demonstrate the limitations of pure vision-centric methods (e.g., SeViLA: 63.6\%) when handling complex reasoning. Among LLM-driven approaches, CAVIA's closed-loop design consistently outperforms open-loop methods like LLoVi and VideoAgent, confirming that iterative refinement and reasoning-guided perception are crucial for comprehensive video understanding. The performance gap is particularly pronounced on IntentQA, where understanding human intentions requires both visual details and sophisticated reasoning—precisely what our dynamic coordination enables.

\begin{figure}[t]
  \centering
  \includegraphics[width=0.9\linewidth]{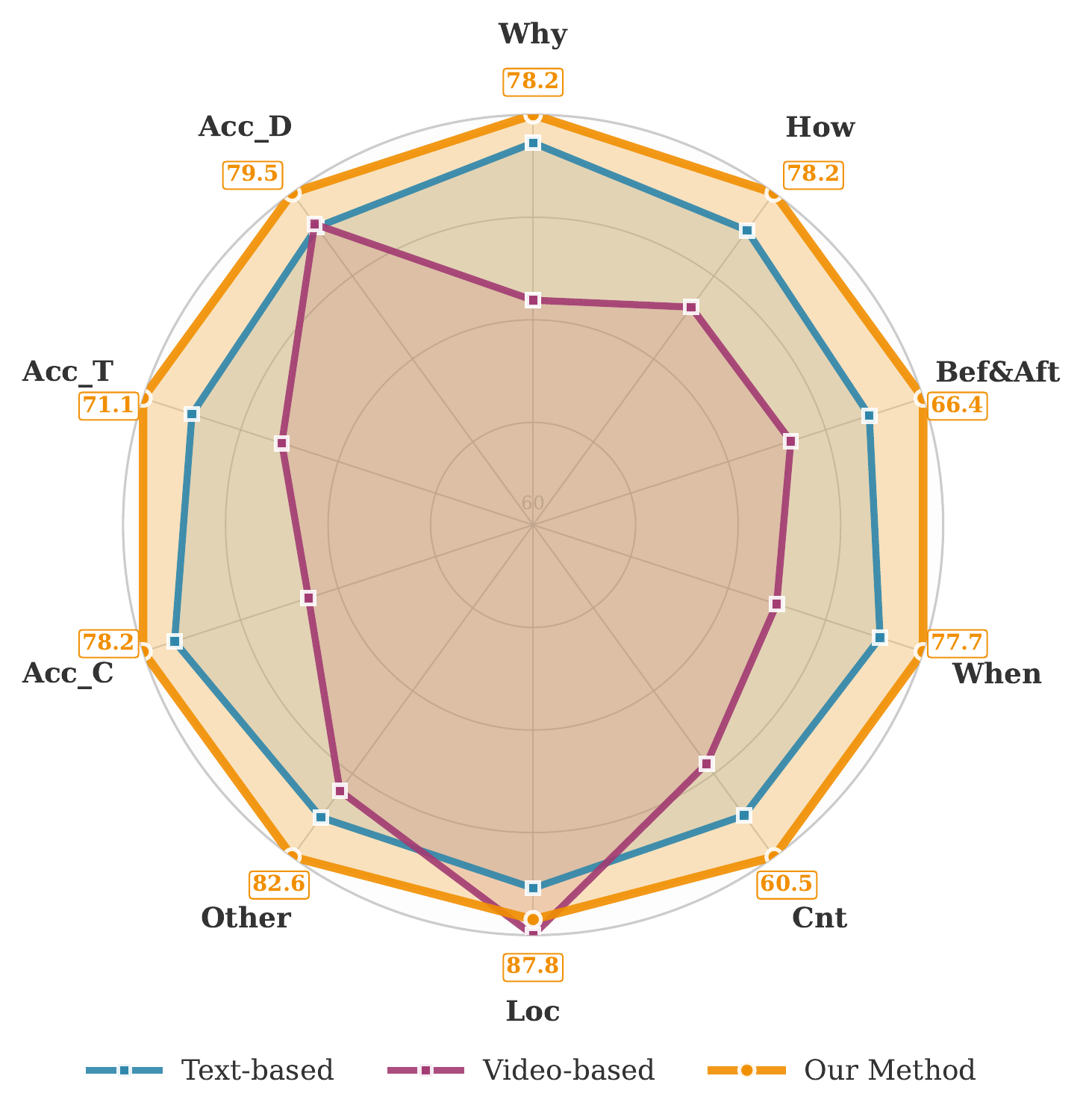} 
      \caption{Performance comparison of text-based methods, video-based methods, and CAVIA across multiple NExT-QA reasoning dimensions including temporal (Why, How, When), spatial (Loc, Bef\&Aft), counting (Cnt), and accuracy metrics for different question categories.}
  \label{fig:radar_chart}
\end{figure}

\subsubsection{Ablation Study: Dissecting the Impact of Core Components}
To validate the effectiveness of our closed-loop architecture, we conduct comprehensive ablation studies examining how each component contributes to overall performance.

\paragraph{Multimodal Coordination Analysis} 
Figure \ref{fig:radar_chart} visualizes the performance comparison across different modalities on various reasoning dimensions. We evaluate text-only baselines (leveraging only caption information), video-only approaches (using VLM without iterative refinement), and our complete CAVIA framework across temporal reasoning (Why, How, When), spatial understanding (Loc, Bef\&Aft), counting tasks (Cnt), and accuracy metrics for different question categories.

The results reveal striking patterns that validate our design choices. Text-based methods excel at semantic reasoning (Acc\_D: 79.5\%) but struggle with fine-grained spatial-temporal understanding, particularly in localization tasks (60.5\%). Conversely, video-based approaches demonstrate strong spatial awareness (Loc: 88.81\%) but fail to maintain consistent performance across reasoning-intensive categories. CAVIA achieves balanced superiority across all dimensions, with particularly notable improvements in temporal reasoning (Why: 78.2\%, How: 78.2\%, When: 77.7\%) where the synergy between iterative caption enhancement and targeted visual extraction proves most valuable. This comprehensive advantage confirms that our closed-loop coordination successfully bridges the complementary strengths of both modalities.

\begin{table}[ht]
\centering
\renewcommand{\arraystretch}{1} 
\setlength{\tabcolsep}{4pt} 
\begin{tabular}{@{}l S[table-format=2.1] S[table-format=2.1]@{}}
\toprule
\textbf{Method} & \textbf{NExT-QA} & \textbf{IntentQA} \\ 
\midrule
\multicolumn{3}{@{}l}{\textbf{Baseline Methods}} \\
LLoVi\textsuperscript{$\alpha$} & 67.7 & 64.0 \\
VideoTree\textsuperscript {$\beta$}& 73.5 & 66.9 \\
\midrule
\multicolumn{3}{@{}l}{\textbf{Efficient LLMs}} \\
COT (Llama-3.1-8B) & 55.7 & 41.5 \\
CAVIA (Llama-3.1-8B) & 61.2 & 53.8 \\
\textbf{Gain} & \textbf{+5.5} & \textbf{+12.3} \\
\addlinespace
COT (Llama-4-Scout) & 64.7 & 62.6 \\
CAVIA (Llama-4-Scout) & 70.6 & 67.9 \\
\textbf{Gain} & \textbf{+5.9} & \textbf{+5.3} \\
\midrule
\multicolumn{3}{@{}l}{\textbf{Large-Scale LLMs}} \\
COT (DeepSeek-V3) & 71.8 & 67.3 \\
CAVIA (DeepSeek-V3) & 74.9 & 69.1 \\
\textbf{Gain} & \textbf{+3.1} & \textbf{+1.8} \\
\addlinespace
COT (GPT-4) & 70.4 & 64.9 \\
CAVIA (GPT-4) & 74.5 & 72.8 \\
\textbf{Gain} & \textbf{+4.1} & \textbf{+7.9} \\
\addlinespace
COT (GPT-4.1) & 71.8 & 65.6 \\
CAVIA (GPT-4.1) & 76.1 & 73.8 \\
\textbf{Gain} & \textbf{+4.3} & \textbf{+8.2} \\
\bottomrule
\end{tabular}
\caption{Ablation study examining CAVIA's performance gains across different base LLMs. We compare our framework against caption-only chain-of-thought (COT) baselines using models ranging from Llama-3.1-8B to GPT-4.1. $\alpha$~\cite{zhang2024simple}, $\beta$~\cite{wang2025videotree}.}
\label{tab:performance}
\end{table}

\paragraph{Impact of Dynamic Prompt Generation Across LLM Scales} 
Table \ref{tab:performance} examines how CAVIA's benefits scale across different base LLMs, from efficient models like Llama-3.1-8B \cite{vavekanand2024llama} to state-of-the-art systems including Llama-4-Scout \cite{meta_llama4_blog} (a MoE architecture with selective parameter activation), DeepSeek-V3 \cite{liu2024deepseek}, GPT-4 and GPT-4.1\footnote{Both GPT-4 and GPT-4.1 can be accessed at \url{https://openai.com/}}.

CAVIA consistently outperforms caption-only COT baselines across all model scales. The improvements are particularly substantial with smaller models (Llama-3.1-8B: +5.5\% on NExT-QA, +12.3\% on IntentQA), where targeted multimodal prompting compensates for limited reasoning capacity. Even with powerful LLMs, CAVIA maintains significant gains (GPT-4.1: +4.3\% on NExT-QA, +8.2\% on IntentQA), demonstrating that our iterative refinement addresses a fundamental limitation: the inability to dynamically request missing visual details during reasoning.

The stronger gains on IntentQA across all models highlight how intent understanding critically benefits from query-specific visual extraction, as inferring human motivations requires nuanced integration of temporal dynamics and spatial context that static captions cannot capture. These model-agnostic improvements validate that closed-loop reasoning-perception coordination represents a fundamental advancement in video understanding methodology, offering a principled solution that enhances any language model's ability to reason about visual content.

\begin{figure}[t]
  \centering
  \includegraphics[width=0.9\linewidth]{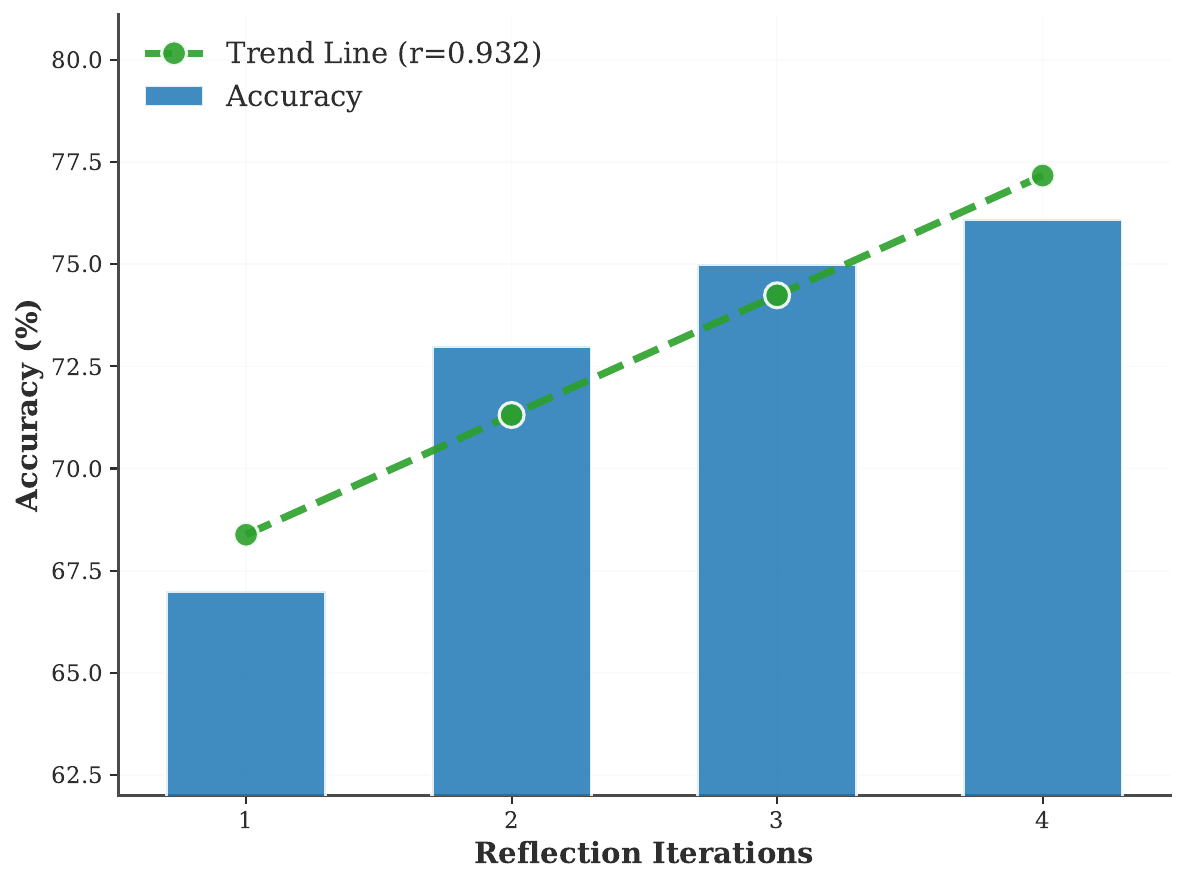} 
    \caption{Accuracy as a function of refinement iterations on NExT-QA dataset. The plot shows performance changes from iteration 1 to 4, with a trend line indicating correlation (r=0.932).}
  \label{fig:accuracy_trend_unified}
\end{figure}

\paragraph{Impact of Iterative Refinement} 
Figure \ref{fig:accuracy_trend_unified} demonstrates the effectiveness of our iterative refinement mechanism. The strong positive correlation (r=0.932) between iteration count and accuracy validates our closed-loop design: performance steadily improves from 67\% (single iteration) to 76\% (four iterations), representing a substantial 9\% absolute gain.

This consistent improvement pattern reveals how each refinement cycle serves a distinct purpose. Early iterations typically identify and address coarse-grained temporal gaps, while later cycles focus on fine-grained spatial details and causal relationships. During each iteration, the LLM identifies specific missing information and generates targeted prompts (e.g., "describe the tool manipulation sequence in frames 45-52"), ensuring that newly extracted information directly addresses reasoning gaps rather than adding redundant details. This targeted approach explains the sustained performance gains across all refinement stages.

\subsection{Efficiency Analysis}
\begin{figure}[t]
  \centering
  \includegraphics[width=0.9\linewidth]{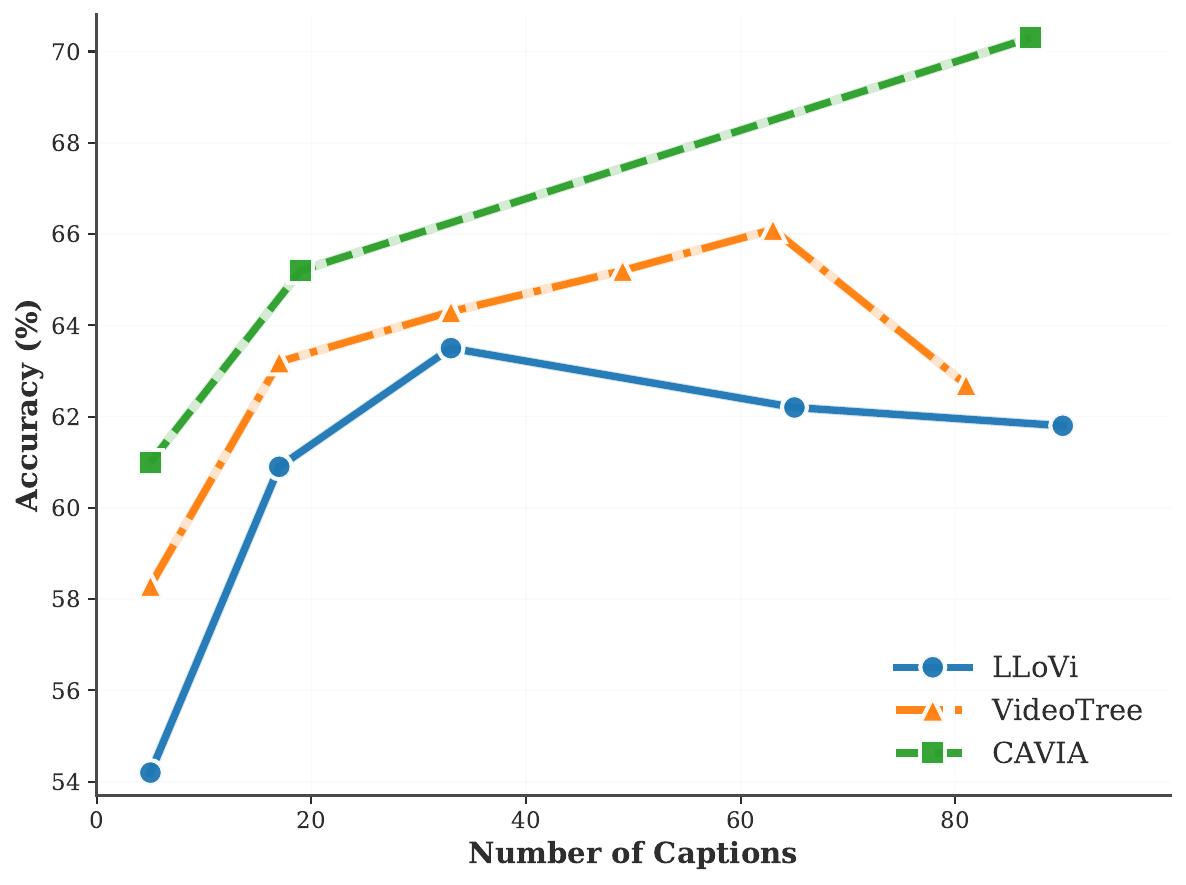} 
    \caption{Accuracy versus number of captions processed on EgoSchema Subset. The plot compares CAVIA with LLoVi and VideoTree across different computational budgets.}
  \label{fig:dual_line_chart}
\end{figure}

Figure \ref{fig:dual_line_chart} illustrates CAVIA's efficiency characteristics by examining the relationship between computational cost (number of captions processed) and performance on EgoSchema Subset. Our framework demonstrates strong performance across the entire computational spectrum, maintaining a consistently superior position in the efficiency-effectiveness space.

This advantage stems from two key design choices. First, our hierarchical localization strategy prioritizes the most informative segments through coarse-to-fine filtering, maximizing early-stage performance gains. Second, the closed-loop architecture ensures intelligent resource utilization: the LLM evaluates information sufficiency at each step and generates targeted prompts only when additional details would improve reasoning, preventing redundant processing while ensuring comprehensive understanding.

CAVIA's stable scaling behavior and flexible computational profile make it particularly suitable for real-world deployment. Whether operating under tight resource constraints or with abundant computational budget, the framework maintains effectiveness through adaptive information extraction. This allows practitioners to dynamically adjust the accuracy-computation trade-off based on specific application requirements, providing a practical solution for diverse video understanding scenarios.

\section{Conclusion}
We presented CAVIA, a closed-loop framework that fundamentally transforms video understanding through dynamic reasoning-perception coordination. By establishing iterative feedback between LLMs and VLMs, CAVIA addresses the critical limitation of static pipelines: the inability to adapt visual processing to query-specific needs. Our hierarchical localization and targeted multimodal prompting enable the system to progressively identify and extract precisely the visual information required for each unique query, transforming video understanding from exhaustive computation to intelligent, adaptive extraction. Extensive experiments across multiple benchmarks validate CAVIA's effectiveness and efficiency, demonstrating that dynamic coordination between reasoning and perception represents a fundamental advancement in video understanding methodology with strong potential for real-world applications.

\section*{Limitations}
While CAVIA demonstrates strong performance across multiple benchmarks, several limitations merit discussion. First, our iterative refinement mechanism increases computational overhead compared to single-pass methods, particularly when multiple iterations are required for complex queries. Although we show efficiency gains through intelligent caption selection, the cumulative cost of repeated LLM-VLM interactions may limit deployment in latency-sensitive applications. Second, CAVIA's effectiveness relies on the quality of initial captions—videos with poor textual descriptions or highly ambiguous visual content may not benefit fully from our dynamic prompting approach. Third, our current implementation uses fixed confidence thresholds for terminating iterations, which may not be optimal across all query types and video domains. Future work could explore adaptive termination criteria and more efficient architectures for iterative reasoning to address these limitations while maintaining the benefits of closed-loop coordination.
\bibliography{custom}

\begin{thebibliography}{42}
\providecommand{\natexlab}[1]{#1}

\bibitem[{Bai et~al.(2025)Bai, Chen, Liu, Wang, Ge, Song, Dang, Wang, Wang, Tang et~al.}]{bai2025qwen2}
Shuai Bai, Keqin Chen, Xuejing Liu, Jialin Wang, Wenbin Ge, Sibo Song, Kai Dang, Peng Wang, Shijie Wang, Jun Tang, and 1 others. 2025.
\newblock Qwen2. 5-vl technical report.
\newblock \emph{arXiv preprint arXiv:2502.13923}.

\bibitem[{Bai et~al.(2023)Bai, Wang, and Chen}]{bai2023glance}
Ziyi Bai, Ruiping Wang, and Xilin Chen. 2023.
\newblock Glance and focus: Memory prompting for multi-event video question answering.
\newblock In \emph{Advances in Neural Information Processing Systems}, volume~36.

\bibitem[{Bala{\v{z}}evi{\'c} et~al.(2024)Bala{\v{z}}evi{\'c}, Shi, Papalampidi, Chaabouni, Koppula, and H{\'e}naff}]{balavzevic2024memory}
Ivana Bala{\v{z}}evi{\'c}, Yuge Shi, Pinelopi Papalampidi, Rahma Chaabouni, Skanda Koppula, and Olivier~J H{\'e}naff. 2024.
\newblock Memory consolidation enables long-context video understanding.
\newblock \emph{arXiv preprint arXiv:2402.05861}.

\bibitem[{Chen et~al.(2024)Chen, Wang, Yan, and Li}]{chen2024spatio}
Zhengxuan Chen, Shuo Wang, Deyang Yan, and Yushi Li. 2024.
\newblock A spatio-temporl deepfake video detection method based on timesformer-cnn.
\newblock In \emph{2024 Third International Conference on Distributed Computing and Electrical Circuits and Electronics (ICDCECE)}, pages 1--6. IEEE.

\bibitem[{Dong et~al.(2025)Dong, Peng, Ma, Wang, Dong, Hu, and Wang}]{dong2025leadqallmdrivencontextawaretemporal}
Xinxin Dong, Baoyun Peng, Haokai Ma, Yufei Wang, Zixuan Dong, Fei Hu, and Xiaodong Wang. 2025.
\newblock \href {https://arxiv.org/abs/2507.14784} {Leadqa: Llm-driven context-aware temporal grounding for video question answering}.
\newblock \emph{Preprint}, arXiv:2507.14784.

\bibitem[{Fei et~al.(2024)Fei, Wu, Ji, Zhang, Zhang, Lee, and Hsu}]{fei2024video}
Hao Fei, Shengqiong Wu, Wei Ji, Hanwang Zhang, Meishan Zhang, Mong-Li Lee, and Wynne Hsu. 2024.
\newblock Video-of-thought: Step-by-step video reasoning from perception to cognition.
\newblock In \emph{International Conference on Machine Learning (ICML)}.

\bibitem[{Gao et~al.(2023)Gao, Zhou, Ji, Zhu, Yang, and Shou}]{gao2023mist}
Difei Gao, Luowei Zhou, Lei Ji, Linchao Zhu, Yi~Yang, and Mike~Zheng Shou. 2023.
\newblock Mist: Multi-modal iterative spatial-temporal transformer for long-form video question answering.
\newblock In \emph{Proceedings of the IEEE/CVF Conference on Computer Vision and Pattern Recognition}, pages 14773--14783.

\bibitem[{Jianqiao et~al.(2025)Jianqiao, Yudi, Hao, Ziheng, Zequn, Zhengjue, Chunhui~Qu, and Xin}]{snapcap2025}
Sun Jianqiao, Su~Yudi, Zhang Hao, Cheng Ziheng, Zeng Zequn, Wang Zhengjue, Chen Chunhui~Qu, Bo, and Yuan Xin. 2025.
\newblock \href {https://nv.opticsjournal.net/J/AI/Issue/2025/1.html} {Snapcap: efficient snapshot compressive scene captioning}.
\newblock \emph{Advanced Imaging}, 2(1):011003.
\newblock Accessed: 2025-07-22.

\bibitem[{Jin et~al.(2024)Jin, Li, Liu, Gu, Wu, Jiang, He, Zhao, Tan, Gan et~al.}]{jin2024efficient}
Yizhang Jin, Jian Li, Yexin Liu, Tianjun Gu, Kai Wu, Zhengkai Jiang, Muyang He, Bo~Zhao, Xin Tan, Zhenye Gan, and 1 others. 2024.
\newblock Efficient multimodal large language models: A survey.
\newblock \emph{arXiv preprint arXiv:2405.10739}.

\bibitem[{Ko et~al.(2023{\natexlab{a}})Ko, Lee, Kang, Roh, and Kim}]{ko2023large}
Dohwan Ko, Ji~Lee, Woo-Young Kang, Byungseok Roh, and Hyunwoo Kim. 2023{\natexlab{a}}.
\newblock Large language models are temporal and causal reasoners for video question answering.
\newblock In \emph{Proceedings of the 2023 Conference on Empirical Methods in Natural Language Processing}, pages 4300--4316.

\bibitem[{Ko et~al.(2023{\natexlab{b}})Ko, Lee, Kang, Roh, and Kim}]{ko2023llama}
Dohwan Ko, Ji~Lee, Woo-Young Kang, Byungseok Roh, and Hyunwoo Kim. 2023{\natexlab{b}}.
\newblock Large language models are temporal and causal reasoners for video question answering.
\newblock In \emph{Proceedings of the 2023 Conference on Empirical Methods in Natural Language Processing}, pages 4300--4316.

\bibitem[{Li et~al.(2023{\natexlab{a}})Li, Wei, Han, and Fan}]{li2023intentqa}
Jiapeng Li, Ping Wei, Wenjuan Han, and Lifeng Fan. 2023{\natexlab{a}}.
\newblock Intentqa: Context-aware video intent reasoning.
\newblock In \emph{Proceedings of the IEEE/CVF international conference on computer vision}, pages 11963--11974.

\bibitem[{Li et~al.(2023{\natexlab{b}})Li, Xiao, Feng, Wang, and Chua}]{li2023discovering}
Yicong Li, Junbin Xiao, Chun Feng, Xiang Wang, and Tat-Seng Chua. 2023{\natexlab{b}}.
\newblock Discovering spatio-temporal rationales for video question answering.
\newblock In \emph{Proceedings of the IEEE/CVF International Conference on Computer Vision}, pages 13869--13878.

\bibitem[{Liao et~al.(2024)Liao, Erler, Wang, Zhai, Zhang, Ma, and Tresp}]{liao2024videoinsta}
Ruotong Liao, Max Erler, Huiyu Wang, Guangyao Zhai, Gengyuan Zhang, Yunpu Ma, and Volker Tresp. 2024.
\newblock Videoinsta: Zero-shot long video understanding via informative spatial-temporal reasoning with llms.
\newblock \emph{arXiv preprint arXiv:2409.20365}.

\bibitem[{Liu et~al.(2025)Liu, Feng, Xue, Wang, Wu, Lu, Zhao, Deng, Zhang, Ruan et~al.}]{liu2024deepseek}
Aixin Liu, Bei Feng, Bing Xue, Bingxuan Wang, Bochao Wu, Chengda Lu, Chenggang Zhao, Chengqi Deng, Chenyu Zhang, Chong Ruan, and 1 others. 2025.
\newblock Deepseek-v3 technical report.
\newblock \emph{arXiv preprint arXiv:2412.19437}.

\bibitem[{Ma et~al.(2024)Ma, Wang, Kong, Wang, Liu, Pei, and Zhao}]{10438044}
Jie Ma, Pinghui Wang, Dechen Kong, Zewei Wang, Jun Liu, Hongbin Pei, and Junzhou Zhao. 2024.
\newblock \href {https://doi.org/10.1109/TPAMI.2024.3366154} {Robust visual question answering: Datasets, methods, and future challenges}.
\newblock \emph{IEEE Transactions on Pattern Analysis and Machine Intelligence}, 46(8):5575--5594.

\bibitem[{Ma et~al.(2025)Ma, Gou, Shi, Sun, Li, Rezatofighi, and Cai}]{ma2025drvideo}
Ziyu Ma, Chenhui Gou, Hengcan Shi, Bin Sun, Shutao Li, Hamid Rezatofighi, and Jianfei Cai. 2025.
\newblock Drvideo: Document retrieval based long video understanding.
\newblock In \emph{Proceedings of the Computer Vision and Pattern Recognition Conference}, pages 18936--18946.

\bibitem[{Mangalam et~al.(2023)Mangalam, Akshulakov, and Malik}]{mangalam2023egoschema}
Karttikeya Mangalam, Raiymbek Akshulakov, and Jitendra Malik. 2023.
\newblock Egoschema: A diagnostic benchmark for very long-form video language understanding.
\newblock \emph{Advances in Neural Information Processing Systems}, 36:46212--46244.

\bibitem[{{Meta AI}(2025)}]{meta_llama4_blog}
{Meta AI}. 2025.
\newblock The llama 4 herd: The beginning of a new era of natively multimodal ai innovation.
\newblock \url{https://ai.meta.com/blog/llama-4-multimodal-intelligence/ }.
\newblock Accessed: 2025-04-05.

\bibitem[{Min et~al.(2024)Min, Buch, Nagrani, Cho, and Schmid}]{min2024morevqa}
Juhong Min, Shyamal Buch, Arsha Nagrani, Minsu Cho, and Cordelia Schmid. 2024.
\newblock Morevqa: Exploring modular reasoning models for video question answering.
\newblock In \emph{Proceedings of the IEEE/CVF Conference on Computer Vision and Pattern Recognition}, pages 13235--13245.

\bibitem[{Nguyen et~al.(2024)Nguyen, Hu, Wu, Nguyen, Ng, and Luu}]{nguyen2024encoding}
Thong~Thanh Nguyen, Zhiyuan Hu, Xiaobao Wu, Cong-Duy~T Nguyen, See-Kiong Ng, and Anh~Tuan Luu. 2024.
\newblock Encoding and controlling global semantics for long-form video question answering.
\newblock In \emph{Proceedings of the 2024 Conference on Empirical Methods in Natural Language Processing}, pages 7109--7126.

\bibitem[{Papalampidi et~al.(2024)Papalampidi, Koppula, Pathak, Chiu, Heyward, Patraucean, Shen, Miech, Zisserman, and Nematzdeh}]{papalampidi2024simple}
Pinelopi Papalampidi, Skanda Koppula, Shreya Pathak, Justin Chiu, Joe Heyward, Viorica Patraucean, Jiajun Shen, Antoine Miech, Andrew Zisserman, and Aida Nematzdeh. 2024.
\newblock A simple recipe for contrastively pre-training video-first encoders beyond 16 frames.
\newblock In \emph{Proceedings of the IEEE/CVF Conference on Computer Vision and Pattern Recognition}, pages 14386--14397.

\bibitem[{Ranasinghe et~al.(2025)Ranasinghe, Li, Kahatapitiya, and Ryoo}]{ranasinghe2024understanding}
Kanchana Ranasinghe, Xiang Li, Kumara Kahatapitiya, and Michael~S Ryoo. 2025.
\newblock Understanding long videos in one multimodal language model pass.
\newblock \emph{arXiv preprint arXiv:2403.16998}, 3.

\bibitem[{Santos et~al.(2025)}]{santos2025infinityvideo}
Saul Santos and 1 others. 2025.
\newblock $\infty$-video: A training-free approach to long video understanding via continuous-time memory consolidation.
\newblock \emph{arXiv preprint arXiv:2501.19098}.

\bibitem[{Sharma and Jalal(2022)}]{sharma2022convolutional}
Himanshu Sharma and Anand~Singh Jalal. 2022.
\newblock Convolutional neural networks-based vqa model.
\newblock In \emph{Proceedings of International Conference on Frontiers in Computing and Systems: COMSYS 2021}, pages 109--116. Springer.

\bibitem[{Shen et~al.(2023)Shen, Gu, Xu, Fan, Wen, and Zhang}]{shen2023accurate}
Yaojie Shen, Xin Gu, Kai Xu, Heng Fan, Longyin Wen, and Libo Zhang. 2023.
\newblock Accurate and fast compressed video captioning.
\newblock In \emph{Proceedings of the IEEE/CVF International Conference on Computer Vision}, pages 15558--15567.

\bibitem[{Song et~al.(2024)Song, Chai, Wang, Zhang, Zhou, Wu, Chi, Guo, Ye, Zhang et~al.}]{song2024moviechat}
Enxin Song, Wenhao Chai, Guanhong Wang, Yucheng Zhang, Haoyang Zhou, Feiyang Wu, Haozhe Chi, Xun Guo, Tian Ye, Yanting Zhang, and 1 others. 2024.
\newblock Moviechat: From dense token to sparse memory for long video understanding.
\newblock In \emph{Proceedings of the IEEE/CVF Conference on Computer Vision and Pattern Recognition}, pages 18221--18232.

\bibitem[{Vavekanand and Sam(2024)}]{vavekanand2024llama}
Raja Vavekanand and Kira Sam. 2024.
\newblock Llama 3.1: An in-depth analysis of the next-generation large language model.

\bibitem[{Wang et~al.(2025{\natexlab{a}})Wang, Zhang, Zohar, and Yeung-Levy}]{10.1007/978-3-031-72989-8_4}
Xiaohan Wang, Yuhui Zhang, Orr Zohar, and Serena Yeung-Levy. 2025{\natexlab{a}}.
\newblock Videoagent: Long-form video understanding with large language model as agent.
\newblock In \emph{Computer Vision -- ECCV 2024}, pages 58--76, Cham. Springer Nature Switzerland.

\bibitem[{Wang et~al.(2024{\natexlab{a}})Wang, Li, Li, Yu, He, Chen, Pei, Zheng, Wang, Shi, Jiang, Li, Xu, Zhang, Huang, Qiao, Wang, and Wang}]{wang2024internvideo2}
Yi~Wang, Kunchang Li, Xinhao Li, Jiashuo Yu, Yinan He, Guo Chen, Baoqi Pei, Rongkun Zheng, Zun Wang, Yansong Shi, Tianxiang Jiang, Songze Li, Jilan Xu, Hongjie Zhang, Yifei Huang, Yu~Qiao, Yali Wang, and Limin Wang. 2024{\natexlab{a}}.
\newblock \href {https://doi.org/10.1007/978-3-031-73013-9_23} {Internvideo2: Scaling foundation models for multimodal video understanding}.
\newblock In \emph{European Conference on Computer Vision (ECCV)}, volume 15143 of \emph{Lecture Notes in Computer Science}, pages 396--416. Springer.

\bibitem[{Wang et~al.(2024{\natexlab{b}})Wang, Wang, Chen, and Zhao}]{wang2024stair}
Yueqian Wang, Yuxuan Wang, Kai Chen, and Dongyan Zhao. 2024{\natexlab{b}}.
\newblock Stair: Spatial-temporal reasoning with auditable intermediate results for video question answering.
\newblock In \emph{Proceedings of the AAAI Conference on Artificial Intelligence}, volume~38, pages 19215--19223.

\bibitem[{Wang et~al.(2025{\natexlab{b}})Wang, Peng, Dong, Fu, Dong, Hu, and Wang}]{10.1007/978-981-96-9961-2_17}
Yufei Wang, Baoyun Peng, Zixuan Dong, Jia Fu, Xinxin Dong, Fei Hu, and Xiaodong Wang. 2025{\natexlab{b}}.
\newblock Optigqa: Llm-driven query optimization for efficient visual grounding in adaptive video question answering.
\newblock In \emph{Advanced Intelligent Computing Technology and Applications}, pages 200--211, Singapore. Springer Nature Singapore.

\bibitem[{Wang et~al.(2025{\natexlab{c}})Wang, Yu, Stengel-Eskin, Yoon, Cheng, Bertasius, and Bansal}]{wang2025videotree}
Ziyang Wang, Shoubin Yu, Elias Stengel-Eskin, Jaehong Yoon, Feng Cheng, Gedas Bertasius, and Mohit Bansal. 2025{\natexlab{c}}.
\newblock Videotree: Adaptive tree-based video representation for llm reasoning on long videos.
\newblock In \emph{Proceedings of the Computer Vision and Pattern Recognition Conference}, pages 3272--3283.

\bibitem[{Wu et~al.(2024)Wu, Li, Chen, and Li}]{wu2024longvideobench}
Haoning Wu, Dongxu Li, Bei Chen, and Junnan Li. 2024.
\newblock Longvideobench: A benchmark for long-context interleaved video-language understanding.
\newblock In \emph{Advances in Neural Information Processing Systems}, volume~37.

\bibitem[{Xiao et~al.(2021)Xiao, Shang, Yao, and Chua}]{xiao2021next}
Junbin Xiao, Xindi Shang, Angela Yao, and Tat-Seng Chua. 2021.
\newblock Next-qa: Next phase of question-answering to explaining temporal actions.
\newblock In \emph{Proceedings of the IEEE/CVF conference on computer vision and pattern recognition}, pages 9777--9786.

\bibitem[{Xiao et~al.(2022)Xiao, Zhou, Chua, and Yan}]{xiao2022video}
Junbin Xiao, Pan Zhou, Tat-Seng Chua, and Shuicheng Yan. 2022.
\newblock Video graph transformer for video question answering.
\newblock In \emph{European Conference on Computer Vision}, pages 39--58. Springer.

\bibitem[{Youk et~al.(2024)Youk, Oh, and Kim}]{youk2024fma}
Geunhyuk Youk, Jihyong Oh, and Munchurl Kim. 2024.
\newblock Fma-net: Flow-guided dynamic filtering and iterative feature refinement with multi-attention for joint video super-resolution and deblurring.
\newblock In \emph{Proceedings of the IEEE/CVF Conference on Computer Vision and Pattern Recognition}, pages 44--55.

\bibitem[{Yu et~al.(2023{\natexlab{a}})Yu, Cho, Yadav, and Bansal}]{yu2023sevila}
Shoubin Yu, Jaemin Cho, Prateek Yadav, and Mohit Bansal. 2023{\natexlab{a}}.
\newblock Self-chained image-language model for video localization and question answering.
\newblock In \emph{Advances in Neural Information Processing Systems}, volume~36.

\bibitem[{Yu et~al.(2023{\natexlab{b}})Yu, Cho, Yadav, and Bansal}]{yu2023self}
Shoubin Yu, Jaemin Cho, Prateek Yadav, and Mohit Bansal. 2023{\natexlab{b}}.
\newblock Self-chained image-language model for video localization and question answering.
\newblock \emph{Advances in Neural Information Processing Systems}, 36:76749--76771.

\bibitem[{Zhang et~al.(2024{\natexlab{a}})Zhang, Lu, Islam, Wang, Yu, Bansal, and Bertasius}]{zhang2024simple}
Ce~Zhang, Taixi Lu, Md~Mohaiminul Islam, Ziyang Wang, Shoubin Yu, Mohit Bansal, and Gedas Bertasius. 2024{\natexlab{a}}.
\newblock A simple llm framework for long-range video question-answering.
\newblock \emph{Proceedings of the Conference on Empirical Methods in Natural Language Processing}, pages 21715--21737.

\bibitem[{Zhang et~al.(2024{\natexlab{b}})Zhang, Wang, Tang, Liu, Feng, Dai, and Jin}]{zhang2024flash}
Haoji Zhang, Yiqin Wang, Yansong Tang, Yong Liu, Jiashi Feng, Jifeng Dai, and Xiaojie Jin. 2024{\natexlab{b}}.
\newblock Flash-vstream: Memory-based real-time understanding for long video streams.
\newblock \emph{arXiv preprint arXiv:2406.08085}.

\bibitem[{Zhang et~al.(2025)Zhang, Yuan, Zhong, Luo, Zhan, Zhang, Hu, and Li}]{zhang2025vl}
Junyang Zhang, Mu~Yuan, Ruiguang Zhong, Puhan Luo, Huiyou Zhan, Ningkang Zhang, Chengchen Hu, and Xiang-Yang Li. 2025.
\newblock A-vl: Adaptive attention for large vision-language models.
\newblock In \emph{Proceedings of the AAAI Conference on Artificial Intelligence}, volume~39, pages 22461--22469.

\end{thebibliography}
\clearpage
\appendix
\section{System Implementation Details}
\subsection{Framework Architecture}

The CAVIA framework implements closed-loop coordination between reasoning and perception through three interconnected modules: Hierarchical Localization for coarse-to-fine filtering, Dynamic Prompting for targeted instruction generation, and Adaptive Extraction for multimodal analysis. Given video $\mathcal{V} = \{f_t\}_{t=1}^T$ with captions $\mathcal{C} = \{c_t\}_{t=1}^T$, we iteratively refine understanding to find optimal answer $a^*$:

\begin{equation}
a^* = \arg\max_{a \in \mathcal{A}} P(a|q, \mathcal{V}, \mathcal{C}^{(k)})
\end{equation}

where $\mathcal{C}^{(k)}$ denotes enhanced captions at iteration $k$.

\subsection{Algorithm Description}

Algorithm \ref{alg:cavia} presents the complete framework with simplified notation for column constraints.

\begin{algorithm}[ht]
\caption{CAVIA Framework}
\label{alg:cavia}
\begin{algorithmic}[1]
\Require Video $\mathcal{V}$, Captions $\mathcal{C}$, Question $q$
\Ensure Answer $a$
\State Init: $\phi \gets 0$, $k \gets 0$, $\mathcal{C}^{(0)} \gets \mathcal{C}$
\State $Q_s \gets$ Decompose($q$)
\While{$\phi < \theta$ \textbf{and} $k < K$}
    \State $B_r \gets$ Retrieve($Q_s$, $\mathcal{C}^{(k)}$)
    \State $G, F_g \gets$ Analyze($q$, $B_r$)
    \If{$|G| > 0$}
        \State $P \gets$ GenPrompts($G$)
        \For{$(p_i, f_i) \in$ zip($P$, $F_g$)}
            \State $d_i \gets$ VLM($\mathcal{V}$, $p_i$, $f_i$)
            \State $\mathcal{C}^{(k+1)}[f_i] \gets$ Merge($\mathcal{C}^{(k)}[f_i]$, $d_i$)
        \EndFor
    \EndIf
    \State $a_L, \phi_L \gets$ LLM($q$, $\mathcal{C}^{(k+1)}$)
    \State $a_V, \phi_V \gets$ VLM($q$, $\mathcal{V}$)
    \State $a, \phi \gets$ Fuse($a_L$, $a_V$, $\phi_L$, $\phi_V$)
    \State $k \gets k + 1$
\EndWhile
\State \Return $a$
\end{algorithmic}
\end{algorithm}

The algorithm operates through five carefully orchestrated phases, each contributing to the closed-loop refinement process:

\textbf{Phase 1 - Query Decomposition}: The Decompose() function transforms complex, multi-faceted queries into atomic sub-questions $Q_s = \{q_1, q_2, ..., q_n\}$. This decomposition targets three primary aspects: temporal relations (e.g., action sequences and timing), spatial configurations (e.g., object locations and movements), and causal dependencies (e.g., action-consequence relationships). By breaking down complex queries, we enable more precise retrieval and focused analysis in subsequent phases.

\textbf{Phase 2 - Hierarchical Retrieval}: The Retrieve() function implements a two-stage filtering mechanism. First, caption sequences are partitioned into semantically coherent blocks $B_i$ of size $w$ (typically 10 frames). Then, relevance scoring is computed as:
\begin{equation}
s(B_i) = \alpha \cdot \text{sim}(q, B_i) + \beta \sum_{q_j \in Q_s} \text{sim}(q_j, B_i)
\end{equation}
where sim() computes cosine similarity in the embedding space, $B_i$ represents caption blocks, and $\alpha$, $\beta$ are weighting parameters balancing global query relevance against sub-question specificity. To ensure high recall in the coarse localization stage, we set the top-K value relatively large (typically $K=5$), preventing potentially relevant segments from being prematurely excluded from the reasoning pipeline during early iterations. This conservative approach proves crucial for complex queries where relevant information may be distributed across multiple non-adjacent segments.

\textbf{Phase 3 - Gap Analysis}: The Analyze() function performs structured examination of retrieved content against query requirements. It outputs a gap set $G = \{g_1, g_2, ..., g_m\}$ where each gap $g_i$ is characterized by: (i) type $\in \{\text{temporal}, \text{spatial}, \text{causal}\}$, (ii) associated frames $F_{g_i} \subset \{1, ..., T\}$, and (iii) specific information requirements. This systematic categorization enables targeted prompt generation in the next phase.

\begin{figure*}[t]
\centering
\includegraphics[width=\textwidth]{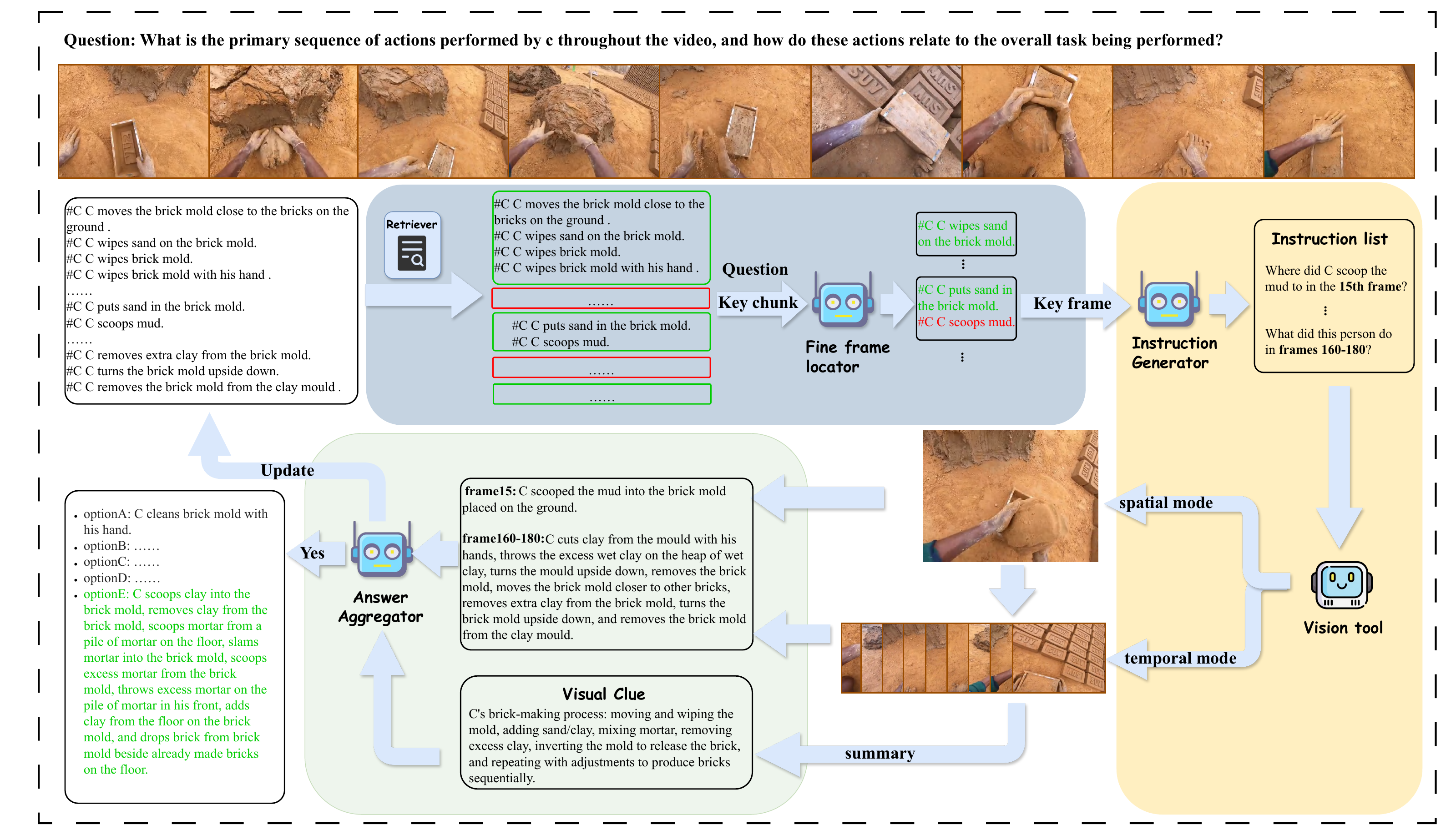}
\caption{CAVIA's iterative refinement on a brick-making task. The system progressively identifies missing clay manipulation details through targeted prompts, demonstrating how closed-loop coordination enables comprehensive understanding of complex sequential activities.}
\label{fig:casestudy}
\end{figure*}

\textbf{Phase 4 - Dynamic Prompt Generation}: Based on identified gaps, GenPrompts() creates specialized instructions $P = \{p_1, p_2, ..., p_m\}$. For temporal gaps, prompts focus on action sequences and state transitions across frame intervals. Spatial prompts direct attention to specific regions or object attributes within individual frames. The prompt generation process adapts to both the gap type and the specific visual content characteristics, ensuring efficient information extraction.

\textbf{Phase 5 - Multimodal Fusion}: The Fuse() function implements adaptive combination of LLM and VLM predictions through confidence-weighted aggregation:
\begin{equation}
a = \arg\max_{a_i \in \mathcal{A}} [\gamma(q) \cdot P_L(a_i) + (1-\gamma(q)) \cdot P_V(a_i)]
\end{equation}
where $\gamma(q) \in [0,1]$ is a query-dependent weighting function learned from question characteristics, $P_L$ and $P_V$ represent probability distributions from LLM and VLM respectively. The weighting adapts based on query type: reasoning-heavy questions favor LLM contributions ($\gamma > 0.5$), while visually-grounded queries emphasize VLM outputs ($\gamma < 0.5$). This adaptive fusion ensures each modality contributes according to its strengths for the specific query at hand.

\section{Experimental Analysis}

\subsection{Case Study: Iterative Refinement Process}
Figure \ref{fig:casestudy} demonstrates CAVIA's iterative process on an EgoSchema brick-making task. Starting with fragmented captions like "C moves brick mold" and "C wipes sand", the system identifies critical gaps in understanding the complete workflow.

In the first iteration, the Retriever identifies relevant caption chunks while the Fine Frame Locator pinpoints frames 15 and 160-180 as containing key information. The gap analysis reveals missing details about tool usage and action sequencing.

The second iteration employs targeted prompting. For spatial understanding, the system queries "Where did C scoop the mud to in the 15th frame?", extracting contextual information about workspace layout. For temporal dynamics, it generates "What did this person do in frames 160-180?", capturing the detailed sequence of clay cutting, mold manipulation, and brick formation.

The final synthesis combines enhanced captions with visual analysis. The spatial mode identifies tools and materials positioning, while temporal mode captures the complete action sequence: scooping mud into molds, removing excess clay, and systematic brick arrangement. This progressive refinement transforms fragmented observations into coherent understanding of the brick-making process.

\subsection{Experimental Environment}

Our experiments utilize a heterogeneous computing environment. Llama-3.1-8B-Instruct and Qwen2.5-VL runs on 2 * NVIDIA RTX 4090 GPU-24GB VRAM each, while Llama-4-Scout requires 4 * NVIDIA A800 GPU-80GB VRAM due to its MoE architecture. All other models including GPT-4, GPT-4.1 and DeepSeek-V3  are accessed through API endpoints. The implementation uses PyTorch 2.1.0 with CUDA 12.1, maintaining consistent hyperparameters across all experiments with confidence threshold 0.85, maximum iterations 5, and temperature settings of 0 for LLMs and VLMs.

\subsection{Ablation Analysis: Component Contributions}

To validate the effectiveness of our closed-loop design, we conduct comprehensive ablation studies examining both individual components and their interactions.

\subsubsection{Component Isolation Analysis}

We systematically remove each component to measure its contribution:
\begin{equation}
\Delta_i = \text{Acc}(\text{Full}) - \text{Acc}(\text{Full} \setminus \{i\})
\end{equation}
where $i$ represents individual components. Beyond quantitative metrics, we analyze qualitative behavioral changes to understand each component's role in the overall system.

\subsubsection{Interaction Effects and Synergy}

The interaction between components reveals interesting patterns. When combining hierarchical localization with dynamic prompting, the joint improvement exceeds the sum of individual contributions:
\begin{equation}
\Delta_{HL+DP} > \Delta_{HL} + \Delta_{DP}
\end{equation}

This synergistic effect stems from the complementary nature of our design: precise localization enables more targeted prompts, while better prompts improve the quality of retrieved information. The closed-loop architecture amplifies these benefits through iterative refinement.

\section{Prompt Engineering}

\subsection{Prompt Design Philosophy}

Our prompt system follows principles of clarity, specificity, and adaptability across five functional categories:

\textbf{Question Decomposition}: Transforms complex queries into atomic sub-questions addressing temporal ("when did X happen?"), spatial ("where is Y located?"), and causal ("why did Z occur?") aspects.

\textbf{Gap Identification}: Structures missing information analysis with explicit categorization, enabling precise targeting of visual extraction efforts.

\textbf{Spatial Extraction}: Directs VLM attention to specific regions with clear attribute specifications, focusing on object identities, positions, and fine-grained features.

\textbf{Temporal Extraction}: Guides sequential analysis across frame intervals, emphasizing state changes, motion patterns, and action continuity.

\textbf{Cross-Modal Synthesis}: Coordinates information fusion between modalities with confidence-based weighting, ensuring balanced contributions based on query characteristics.

\end{document}